\documentclass[10pt,conference]{IEEEtran}
\IEEEoverridecommandlockouts
\usepackage{cite}
\usepackage{amsmath,amssymb,amsfonts}
\usepackage[caption=false]{subfig}
\usepackage{algorithmic}
\usepackage{graphicx}
\usepackage{textcomp}
\usepackage{xcolor,colortbl}
\usepackage{booktabs}
\usepackage{multirow}
\usepackage{mhchem}
\IEEEoverridecommandlockouts
\def\BibTeX{{\rm B\kern-.05em{\sc i\kern-.025em b}\kern-.08em
    T\kern-.1667em\lower.7ex\hbox{E}\kern-.125emX}}
\begin{document}

\title{A Brain-Inspired Deep Separation Network for Single Channel Raman Spectra Unmixing
\thanks{* Corresponding author\\ \hspace*{0.8em} This work was supported by the National Natural Science Foundation of China under Grant 62476144 and 62076140.}
}

\IEEEpubid{
	\makebox[\columnwidth]{
		\parbox{\columnwidth}{
			\vspace{3em} 
			{\scriptsize \copyright 2026 IEEE. Personal use of this material is permitted. Permission from IEEE must be obtained for all other uses, in any current or future media, including reprinting/republishing this material for advertising or promotional purposes, creating new collective works, for resale or redistribution to servers or lists, or reuse of any copyrighted component of this work in other works.}
		}
	} \hspace{\columnsep}\makebox[\columnwidth]{}
}

\author{
    \IEEEauthorblockN{
        Gaoruishu Long\textsuperscript{1}, 
        Jinchao Liu\textsuperscript{1,*}, 
        Bo Liu\textsuperscript{2,3}, 
        Jie Liu\textsuperscript{1}, 
        Xiaolin Hu\textsuperscript{4}
    }
    \IEEEauthorblockA{
        \textsuperscript{1}College of Artificial Intelligence, Nankai University, Tianjin, China \\
        \textsuperscript{2}Guangdong Laboratory of Chemistry and Fine Chemical Industry, Guangdong, China \\
        \textsuperscript{3}Smekal Tech (Shantou) Ltd. \\
        \textsuperscript{4}Department of Computer Science and Technology, Tsinghua University, Beijing, China \\
    }
}

\maketitle

\begin{abstract}
Raman spectra obtained in real world applications are often a noisy combination of several spectra of various substances in a tested sample. Unmixing such spectra into individual components corresponding to each of the substances is of great value and has been a longstanding challenge in Raman spectroscopy.  
Existing unmixing methods are predominantly designed to invert an overdetermined mixed model and therefore require multiple mixed spectra as input. However, open domain and/or non-cooperative detection applications in Raman spectroscopy such as controlled substance detection, call for single-channel solutions which can identify individual components from thousands of candidates by analyzing only a single noisy mixed spectrum. To our knowledge, sparse regression is the only existing solution which can cope with this scenario, yet it has very low tolerance to noises and can hardly be applicable in practice. 
To address these limitations, we introduce a novel neural approach for single-channel Raman spectrum unmixing inspired by speech separation. It aims at solving underdetermined systems and can decompose a noisy mixed spectrum from a library of thousands of components (substances). The core of our method is a deep separation neural network (RSSNet) which takes a mixed spectrum as input and outputs spectra of pure components. 
We created two synthetic datasets of single-channel Raman spectra unmixing and demonstrated feasibility and superiority of RSSNet on these datasets (outperform competing methods by $>$4dB). Furthermore, we verified that RSSNet, trained solely on synthetic data, can successfully unmix real-world mixed spectra of mixtures of mineral powders, exhibiting strong generalization.
Our approach represents a new paradigm for Raman unmixing and enables new possibilities for fast detection of Raman mixtures.
\end{abstract}

\begin{IEEEkeywords}
Blind Source Separation, Brain-Inspired Computing, Raman Spectroscopy, Signal Processing
\end{IEEEkeywords}

\section{Introduction}

Raman spectroscopy serves as a vital non-destructive tool for material identification, relying on the unique “molecular fingerprint” generated by vibrational modes~\cite{liu2017deep}. However, in practical scenarios such as hazardous substance detection or pharmaceutical analysis, the observed spectra are often linear mixtures of multiple components. The objective of Raman unmixing is to blindly decompose a mixed spectrum $\mathbf{y}$ into its constituent pure spectra (\textit{endmembers}) and their corresponding proportions (\textit{abundances}). While effective in controlled settings, this inverse problem becomes non-trivial when the pure components are unknown (i.e., blind source separation).

\begin{table*}[t]
	\centering
    \small
   	\caption{{\small Comparison of unmixing paradigms for Raman spectroscopy and spectral unmixing in general. U.D. stands for underdetermined, O.D. stands for overdetermined.}}
	\begin{tabular}{lcccc}
		\toprule
		\multirow{2}*{\textbf{Unmixing Paradigms}} & \multirow{2}*{\textbf{Methods}}   &\textbf{Mixing}  &\multirow{2}*{\textbf{\#Inputs}} & \textbf{Noise}\\
        & &  \textbf{Model} & &\textbf{Tolerance}\\
		\midrule
        Sparse regression-based &  \hfill e.g.~\cite{Sunsal2010,SunsalLegacy2021}   & U.D. & {\color{green!70!blue}Single}   &  {\color{red}Low}\\        
		\addlinespace[0.5ex]    
        Geometrical and statistical-based & \hfill e.g.~\cite{FLCS2001,garrido2008multivariate,kouakou2024fly}  & O.D. & {\color{red}Multiple} & {\color{red}Low}\\
        \addlinespace[0.5ex]
        Hybrid classical-learning-based & \hfill e.g.~\cite{UnmixingAE2024PNAS,mDAE,SIDAEU,Endnet,CNNAEU} & O.D. & {\color{red}Multiple}   & {\color{green!70!blue}High}\\
		\addlinespace[0.5ex]
        \rowcolor{blue!10} 
        \textbf{Neural separation-based} & \hfill  \textbf{RSSNet (ours)}  & U.D. & {\color{green!70!blue}Single} & {\color{green!70!blue}High} \\
		\bottomrule
	\end{tabular}
	\label{tab_1}
\end{table*}

{\color{black}

Current unmixing paradigms face fundamental limitations in this context, as summarized in Table~\ref{tab_1}. Standard geometrical and statistical approaches (e.g., NMF~\cite{5585746}, VCA~\cite{VCA2005}) rely on statistical redundancy across multiple observations to solve the unmixing problem. Consequently, they fail in single-channel scenarios where the system becomes severely underdetermined. Similarly, while deep learning has shown potential in remote sensing~\cite{mDAE,SIDAEU,Endnet,CNNAEU} and was recently explicitly adapted for Raman hyperspectral imaging~\cite{UnmixingAE2024PNAS}, these methods predominantly rely on spatial correlations and are typically constrained to resolving a limited number of endmembers in overdetermined settings. They performed extremely unsatisfactorily in scenarios involving a vast amount of (possible) pure components where ill-posed underdetermined systems need to be inverted, as we demonstrated in this study. In scenarios restricted to a single spectrum, sparse regression stands as the sole conventional candidate, bypassing the underdetermined limit via pre-defined dictionaries. However, a major drawback is that these methods are extremely sensitive to noise~\cite{Li2017Sparse,XU201846,Gong2017Multiobjective,Drumetz2016Blind}, making them impractical for real-world signals where low signal-to-noise ratios (SNR) degrade decomposition accuracy.

This limitation is critical in non-cooperative applications (e.g., controlled substance detection) where acquiring a full hyperspectral image is infeasible. Consequently, there is an urgent need for \textit{single-channel Raman unmixing—identifying individual components from thousands of candidates using only a single, often noisy, input}. Motivated by blind source separation, speech separation in particular, we propose a neural network approach for single-channel Raman unmixing. Although speech processing deals with temporal audio signals, the fundamental principles of extracting meaningful individual components from complex mixtures are highly transferable to the challenges encountered in spectral analysis. Especially deep learning based speech separation methods have been proposed and showed great potential~\cite{8369155,ConvTasNet,DPRNN2020,TDANet2023}.

Our main contributions are summarized as follows:
\begin{itemize}
    \item We present a novel neural separation-based paradigm inspired by speech separation for single-channel spectral unmixing in Raman spectroscopy.
    \item We present the first neural network architecture, named RSSNet, which can decompose a single noisy Raman spectrum of a mixture from a library of thousands of substances with high accuracy. 
    \item We validated the superiority of our method on both synthetic mineral datasets and collected real-world samples, and the extensive experiments demonstrate that RSSNet outperforms competing baselines by over 4 dB, exhibiting strong generalization where traditional methods fail.
\end{itemize}

The benchmark datasets used in this paper are available at https://github.com/Hydrochlor/RSSNet.

\section{Related Work}

\textbf{Spectral Unmixing Approaches.} 
Existing methods largely fall into three categories. \textit{Sparse regression} (e.g., SUnSAL~\cite{Sunsal2010}) relies on comprehensive spectral libraries to solve the unmixing problem. While applicable to single-spectrum inputs, these methods are highly sensitive to noise and dictionary completeness. \textit{Geometrical and statistical methods} (e.g., NMF~\cite{5585746}, VCA~\cite{VCA2005}) exploit the simplex structure of data but require multiple mixed observations (overdetermined systems) to estimate endmembers blindly, making them unsuitable for single-channel scenarios. Recently, \textit{Hybrid classical-learning-based unmixing} (e.g., Autoencoders~\cite{UnmixingAE2024PNAS}) has emerged, utilizing neural networks to model spectral mixing. However, these models typically rely on spatial constraints within an image cube and struggle to decompose a single independent mixture into a vast number of potential components without spatial context.

\textbf{Single-Channel Speech Separation.} 
Our work draws inspiration from single-channel speech separation, often formulated as the "Cocktail Party Problem." Recent advancements have shifted from time-frequency masking to time-domain separation, yielding state-of-the-art performance. Architectures such as Conv-TasNet~\cite{ConvTasNet} and DPRNN~\cite{DPRNN2020} model raw waveforms directly, effectively handling long-range dependencies. TDANet~\cite{TDANet2023} further improved efficiency using top-down attention mechanisms. These methods have demonstrated strong capabilities in disentangling overlapping signals from a single input, a characteristic we aim to transfer to spectral unmixing.

\section{Method} \label{sec:method}

Suppose that we have a mixed spectrum denoted as $\mathbf{y} \in \mathbb{R} ^ {1 \times L}$, where $L$ represents the length of the mixed spectrum. Under the linear mixing assumption, we have:
\begin{equation}
	\mathbf{y} = \sum_{i = 1}^{C} \alpha_i \mathbf{s}_i + \mathbf{e},
    \label{Eq_linearmix}
\end{equation}
where $\mathbf{s}_i \in \mathbb{R} ^ {1 \times L}$ denotes the $i$-th individual substance spectrum, $\alpha_i$ denotes the proportion of $\mathbf{s}_i$, often referred to as abundance or concentration. $\mathbf{e} \in \mathbb{R} ^ {1 \times L}$ denotes the noise signal. Our goal is to unmix $\mathbf{y}$ by directly inverting Eq.~(\ref{Eq_linearmix}) using a neural network 
\begin{align}
    \{\mathbf{\hat{s}}_i, i = 1, \cdots, C\} = \mathcal{U}_{\phi}(\mathbf{y}),
	\label{Eq_l0minnoisefree}
\end{align}
where $\mathcal{U}$ denotes a neural network with trainable weights $\phi$ to unmix $\mathbf{y}$.

We use the scale-invariant source-to-noise ratio (SI-SNR)~\cite{DeepClustering} to quantitatively measure the difference between the estimated individual spectra $\mathbf{\hat{s}}_i$ and the corresponding ground truth $\mathbf{s}_i$.

\subsection{Overall architecture}

\begin{figure*}
	\centering
	\includegraphics[width=0.7\textwidth]{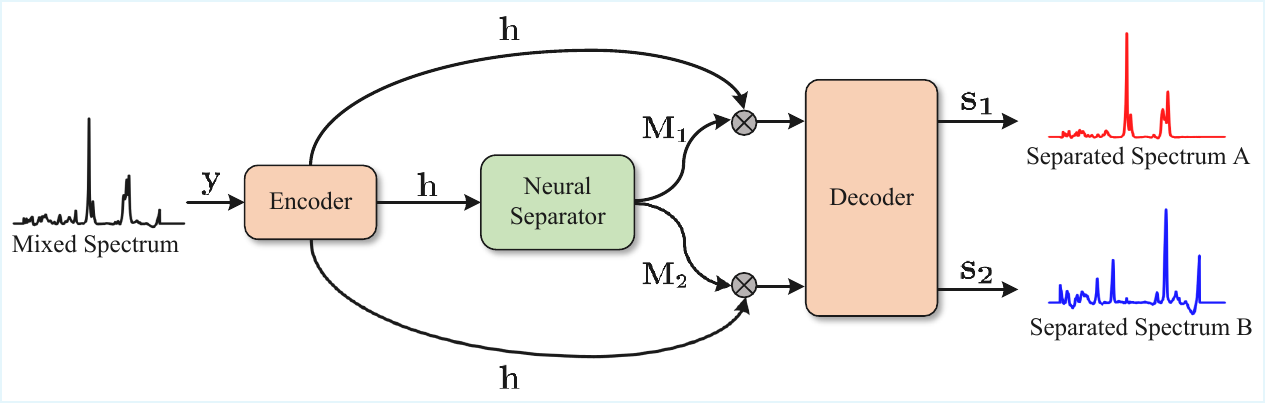}
	\caption{{\small Graphical illustration of single channel Raman spectra unmixing using a deep separation network. $\otimes$ denotes element-wise multiplication. The diagram assumes that there are only two distinct individual substance spectra in the mixed spectrum.}}
	\label{fig:1}
\end{figure*}

In terms of overall pipeline, our method admits the popular encoder-separator-decoder (ESD) framework which has proved to be very effective in speech separation~\cite{TasNet,AFRCNN2021,TDANet2023}, as shown in Fig.~\ref{fig:1}. The hypothesis behind this ESD framework is that it allows learning a feature space where separation can be carried out with a better accuracy and efficiency. 

\subsection{Raman encoder and decoder}

The encoder and decoder form the entrance and exit of our separation pipeline, respectively. Inspired by time-domain speech separation~\cite{TasNet}, we employ a lightweight encoder consisting of a 1-D convolution layer followed by Global Layer Normalization (GLN) and PReLU to project the mixed spectrum $\mathbf{y}$ into a latent feature representation $\mathbf{h} = \text{PReLU}(\text{GLN}(\text{conv1d}(\mathbf{y})))$, where $\mathbf{h} \in \mathbb{R} ^ {N \times L'}$. Symmetrically, the decoder reconstructs the estimated pure spectra $\mathbf{\hat{s}}_i$ from the masked features using a transposed convolution with identical kernel size and stride: $\mathbf{\hat{s}_i} = \text{convtranspose1d}(\mathbf{h} \otimes \mathbf{M_i})$. 

We note that the encoder's kernel size is a critical hyperparameter for capturing sharp spectral peaks, as analyzed in our ablation study (See Table~\ref{tab:ablation}).

\subsection{Neural separator} \label{subsec_separation}

\begin{figure*}[tb]
	\centering
	\includegraphics[width=0.7\textwidth]{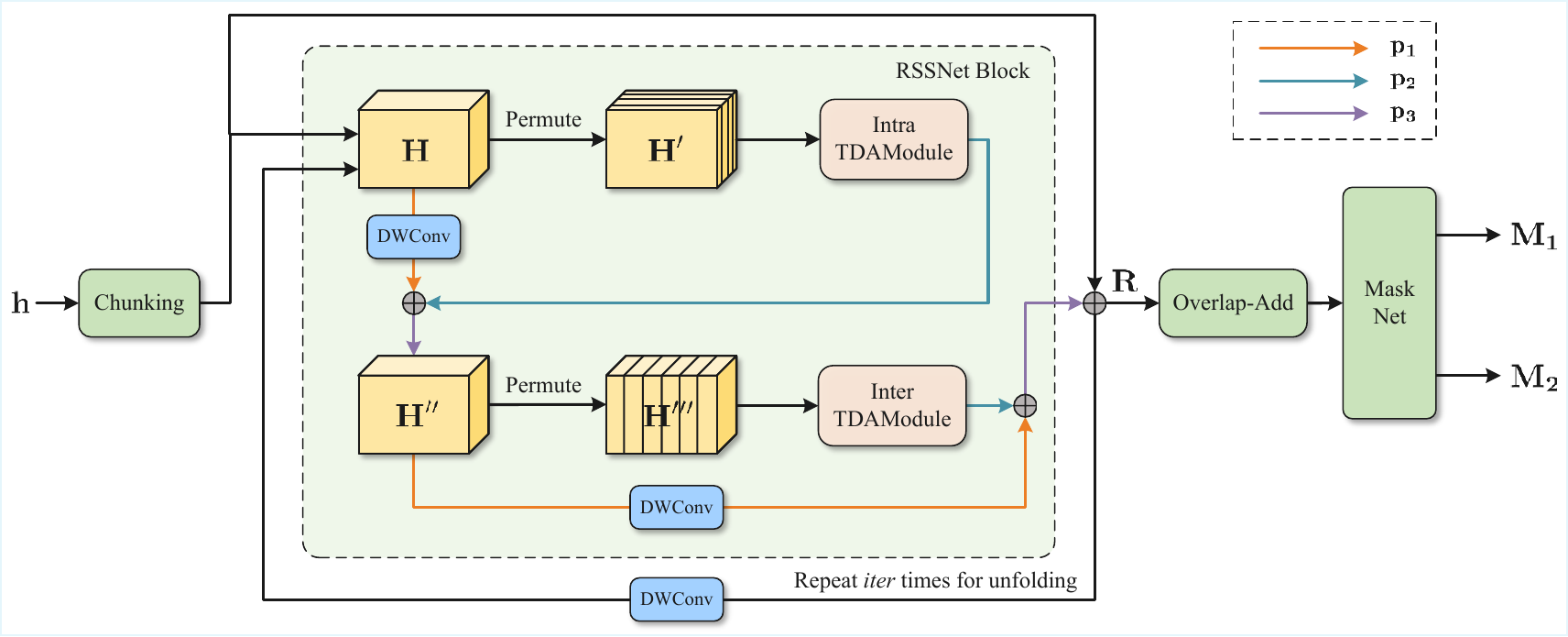}
	\caption{{\small The architecture of our separation network. $\oplus$ denotes element-wise addition, and \textit{DWConv} denotes a depth-wise convolution layer. $\mathbf{p_1}$, $\mathbf{p_2}$ and $\mathbf{p_3}$ denote the paths for ablation study reported in Section \ref{sec:ablation}. Here, separation network still assumes that there are only two distinct individual substance spectra in the mixed spectrum, taking $\mathbf{h}$ as input and finally obtains two masks, $\mathbf{M_1}$ and $\mathbf{M_2}$.}}
	\label{fig:2}
\end{figure*}

As illustrated in Fig.~\ref{fig:2}, the separator follows a dual-path design inspired by speech separation methods~\cite{DPRNN2020,TDANet2023}. It begins with a Chunking stage that segments the input feature into overlapping chunks to facilitate local-global modeling. These chunks are processed by stacked RSSNet Blocks, which incorporate Top-Down Attention (TDA) modules to capture multi-scale spectral dependencies. Subsequently, an Overlap-Add operation reconstructs the sequence, which is finally mapped to component-specific masks via the Mask Net.

\textbf{Chunking.}
The input feature $\mathbf{h}$ is firstly segmented into $T$ overlapping chunks of size $K$ and stride $K / 2$, forming a 3-D tensor $\mathbf{H} \in \mathbb{R} ^ {N \times K \times T}$.

\textbf{RSSNet blocks.}
The core design of the neural separator followed the dual path structure proposed in DPRNN~\cite{DPRNN2020}, within each path, a brain-inspired TDA module was adopted to enhance the ability of learning local and global dependencies. TDA module is a neural network designed with a top-down attention module, which modulates features of different scales top-down through the attention signal obtained from multi-scale features~\cite{TDANet2023}. 
To be specific, when the RSSNet Block first receives the input $\mathbf{H}$, it performs a permute operation on it, reshaping the dimensions of $\mathbf{H}$ to $\mathbf{H'} \in \mathbb{R} ^ {T \times N \times K}$, indicating that it initially processes features along the dimension $T$. An Intra-TDA module then receives H' to learn local dependencies, which refers to the integration of features within the chunk. 

The output of Intra-TDA module is added to $\mathbf{H}$ which passed through a depth-wise convolution layer, resulting in the input $\mathbf{H''}$ for the Inter-TDA module section. Notably, this part is more akin to a two-branch structure rather than a residual path, implying that the network's architecture incorporates two distinct pathways for feature propagation and modulation, differing from the conventional residual paths commonly observed in neural networks. We will further explain its role later. The Intra-TDA module can be represented mathematically as follows: 
\begin{equation}
	\mathbf{H''} = f_{1}(\text{Permute}(\mathbf{H})) + \text{conv2d}(\mathbf{H}),
\end{equation}
where $f_{1}(\text{·})$ denotes the network function fitted by Intra-TDA module.

The structure of the Inter-TDA module section is similar to that of the Intra-TDA module section. The input $\mathbf{H''}$ is first permuted, reshaping it to $\mathbf{H'''} \in \mathbb{R}^{K \times N \times T}$, which means this section processes features along the dimension $K$. Subsequently, it passed through Inter-TDA module, which learns global dependencies, specifically integrating features between chunks. Then, the output is obtained by adding it to $\mathbf{H''}$ passed through a convolution layer, resulting in the final output of the RSSNet Block. Similarly, the mathematical expression can be written as:
\begin{equation}
	\mathbf{H''''} = f_{2}(\text{Permute}(\mathbf{H''})) + \text{conv2d}(\mathbf{H''}),
\end{equation}
where $\mathbf{H''''}$ denotes the final output of the block, $f_2(\text{·})$ denotes the network function fitted by Inter-TDA module.

Note that the depth-wise convolution layer on the path $p_{1}$ improved the performance (Table \ref{tab:ablation}). Intuitively speaking, learning local dependencies occurs along the dimension $T$, and global along the dimension $K$. Our depth-wise convolution layer \textit{DWConv} here, due to its characteristic of performing separate convolution operations on each input channel, can be regarded as processing features along the dimension $N$. The \textit{DWConv} path, based on the TCN concept, emulates the self-attention module in the transformer~\cite{donghao2024moderntcn}. It allows for further extraction of relationships between feature sequences within each channel on top of local and global dependencies. As a result, it enhances the network's ability to model global correlations in spectral sequence.

Similarly to TDANet, we also consider the RSSNet Block as a recurrent model, and repeat the entire RSSNet Block $iter$ times. If we denote the input and output of the $i$-th iteration of the RSSNet Block as $\mathbf{R_i}$ and $\mathbf{R'_i}$, we can obtain the output of the $i$-th iteration as follows:
\begin{equation}
	\mathbf{R_{i+1}} = \text{conv2d}(\mathbf{R'_i} + \mathbf{H}).
\end{equation}

Typically, the convolution layer is set with a kernel size of $1 \times 1$. The addition of convolution layer is aimed at further integrating features from different scales. The output of the overall separation network is $\mathbf{R'_{iter}}$.

\textbf{Intra and inter TDA module.}
The TDA module mimics the top-down modulation in the biological visual and auditory cortex, where high-level semantic information guides the filtering of low-level sensory signals~\cite{TDANet2023}, as illustrated in Fig. \ref{fig:3}. It integrates downsampling, global attention (GA) module, and local attention (LA) module to process sequential data effectively.

Initially, the input $\mathbf{L_0}$ undergoes feature extraction through a series of downsampling layers, progressively reducing the sequence length by half each layer while capturing multi-scale features. These features are aggregated to form global features $\mathbf{G}$ via average pooling and summation, ensuring a holistic representation of the input. The Global Attention (GA) module, inspired by the Transformer architecture~\cite{vaswani2017attention}, refines $\mathbf{G}$ by applying a multi-head attention (MHA) layer and a feed-forward network (FFN) layer, producing global top-down attention $\mathbf{\hat{G}}$. The global top-down attention $\mathbf{\hat{G}}$ is then interpolated and used to modulate the multi-scale features.

The local attention (LA) module gradually restores the modulated features back to the original resolution. By generating adaptive parameters $\mathbf{\rho}$ and $\mathbf{b}$ derived from adjacent feature scales as local top-down attention, the LA module progressively refines the feature representations, reconstructing fine-grained features.

\begin{figure}[!t]
	\centering
	\subfloat[]{
		\label{fig:3.1}
		\includegraphics[scale=0.4]{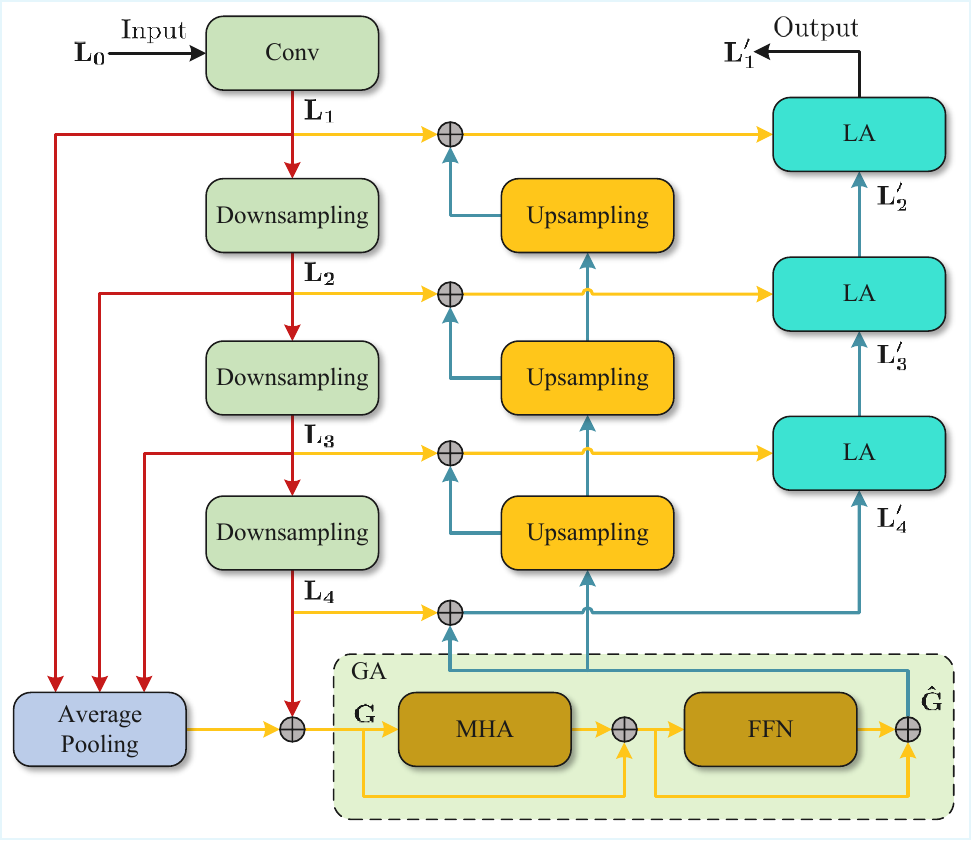}
	}
    \quad
	\subfloat[]{
		\centering
		\label{fig:3.2}
		\includegraphics[scale=0.4]{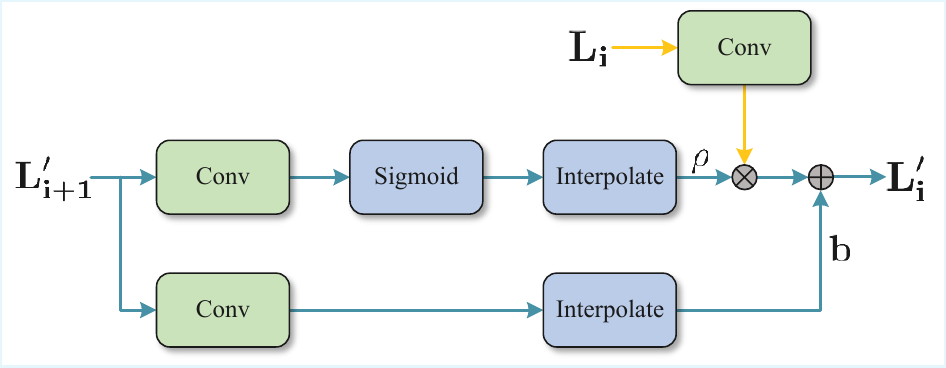}
	}
	\caption{\textbf{(a)} The architecture of the TDA module. $\oplus$ denotes element-wise addition. Here, we assume that the downsampling depth $S = 3$. The red, yellow, and blue arrows correspond to bottom-up connections, lateral connections, and top-down connections, respectively. \textbf{(b)} The internal structure of the LA layer. $\oplus$ denotes element-wise addition, $\otimes$ denotes element-wise product.}
	\label{fig:3}
\end{figure}

\textbf{Overlap-add and mask net.}
Overlap-add is the inverse operation of chunking. Define $\mathbf{R} = \mathbf{R'_{iter}} \in \mathbb{R}^{N \times K \times T}$ as the input to the overlap-add section, it is responsible for transforming the divided chunks back into a feature sequence $\mathbf{\hat{R}} \in \mathbb{R}^{N \times L'}$. Then $\mathbf{\hat{R}}$ is used as the input to Mask Net, which generates masks $\mathbf{M_i} \in \mathbb{R}^{N\times L'}$ for $C$ individual material spectra. Mask Net consists of a PReLU layer followed by a 1-D convolution layer with a kernel size of 1.

\section{Experiments} \label{sec:experiments}

In our experiments, we set the feature dimension $N=256$, encoder/decoder kernel size to 3 (stride 1), and chunk size $K = \sqrt{2L} \approx 45$ according to ~\cite{DPRNN2020}. The separator comprises 6 stacked RSSNet blocks. Within the TDA modules, we configure the upsampling depth $S=3$, use 8 attention heads (512-dim), and set the FFN hidden dimension to 1024. All depth-wise convolutions use $1\times1$ kernels and stride of 1, and the dropout rate is fixed at 0.1.

We use Adam as the optimizer with an initial learning rate of 1e-3. Gradient clipping is also applied during training, limiting the gradient's L2 norm to 5. 
Training ran for 100 epochs (\textit{RRUFF-2Mix}) and 200 epochs (\textit{UNIPR-2Mix}) on 2$\times$GeForce RTX 3080 GPUs.

\begin{table*}[tb]
	\centering
    \caption{{\small Comparison of the proposed RSSNet and competing methods for Raman unmixing on the datasets RRUFF-2Mix and UNIPR-2Mix. TF stands for Transformer and NA stands for Not Applicable. }}
    \resizebox{\linewidth}{!}{
	\begin{tabular}{lccccc}
		\toprule
		\multirow{2}*{\textbf{Methods}} & \multicolumn{2}{c}{\textbf{RRUFF-2Mix}} & 
        \multicolumn{2}{c}{\textbf{UNIPR-2Mix}} & \multirow{2}*{\textbf{Params} (M)} \\ \cmidrule{2-3}  \cmidrule{4-5}
		  & SI-SNR (dB) $\uparrow$ & SI-SNRi (dB) $\uparrow$ & SI-SNR (dB) $\uparrow$ & SI-SNRi (dB) $\uparrow$ & \\
		\midrule
        \textbf{Sparse regression-based}&  &  &  &  &  \\
        \addlinespace[0.5ex]
        SUnSAL~\cite{Sunsal2010} & Failed & - & Failed & - & - \\
        NNOMP~\cite{NNOMP_Unmixing} & Failed & - & Failed & - & - \\
        \midrule
        \addlinespace[0.5ex]
        \textbf{Geometrical and statistical-based} & & & & & \\        
        \addlinespace[0.5ex]
        FastICA~\cite{FastICA1997} & -20.18 & - & -22.69 & - & - \\
        NMF~\cite{NMF1999} & -15.72 & - & -22.97 & - & - \\
        VCA~\cite{VCA2005} & -18.41 & - & -22.05 & - & - \\
        N-FINDR~\cite{NFINDR1999} & -17.84 & - & -22.05 & - & - \\
        FCLS~\cite{FLCS2001}      & Failed & - & Failed & - & -\\
        MCR-ALS~\cite{MCRALSraman2022}     & Failed & - & Failed & - & -\\
        KF-OSU~\cite{kouakou2024fly}      & - & - & -27.06 & - & -\\
        \midrule
        \addlinespace[0.5ex]
        \textbf{Hybrid classical-learning-based}&  &  &  &  &  \\
        \addlinespace[0.5ex]
        Unmixing-AE(Conv.)~\cite{UnmixingAE2024PNAS} & -21.33 & - & -22.61 & - & - \\
        Unmixing-AE(Conv. TF)~\cite{UnmixingAE2024PNAS} & -20.56 & - & -22.51 & - & - \\
        Unmixing-AE(Dense)~\cite{UnmixingAE2024PNAS} & -21.67 & - & -22.86 & - & - \\
        Unmixing-AE(TF)~\cite{UnmixingAE2024PNAS} & -20.48 & - & -22.86 & - & - \\
        mDAE~\cite{mDAE}        & -16.79 & -   & -22.04 & - & -  \\
		SIDAEU~\cite{SIDAEU}      & -16.96 & -    & -22.14 & - & -  \\
		CNNAEU~\cite{CNNAEU}     & -17.12 & -    & NA & - & - \\
        Endnet~\cite{Endnet}      & Failed & - & Failed & - & -\\
        \midrule
        \addlinespace[0.5ex]
        \textbf{Neural separation-based} & & & & & \\
        \addlinespace[0.5ex]
        Conv-TasNet~\cite{ConvTasNet}    & 13.75 {\scriptsize $\pm 0.38$} & 15.12 {\scriptsize $\pm 0.38$}  & 11.79 {\scriptsize $\pm 0.12$} & 12.21 {\scriptsize $\pm 0.12$} & 3.46 \\
		DPRNN~\cite{DPRNN2020}   & 16.73 {\scriptsize $\pm 0.53$} & 18.09 {\scriptsize $\pm 0.53$}  & 12.44 {\scriptsize $\pm 0.14$} & 12.85 {\scriptsize $\pm 0.14$} & 38.98 \\
		AFRCNN~\cite{AFRCNN2021}  & 15.90 {\scriptsize $\pm 0.79$} & 17.27 {\scriptsize $\pm 0.79$}  & 12.06 {\scriptsize $\pm 0.11$} & 12.48 {\scriptsize $\pm 0.11$} & 15.73 \\
		TDANet~\cite{TDANet2023}  & 11.74 {\scriptsize $\pm 0.34$} & 13.11 {\scriptsize $\pm 0.34$}  & 9.89 {\scriptsize $\pm 0.19$} & 10.30 {\scriptsize $\pm 0.19$} & \textbf{2.50}\\
		\midrule
        \rowcolor{blue!10} 
		\textbf{RSSNet (ours)} & \textbf{21.13} {\scriptsize $\pm 0.81$} & \textbf{22.50} {\scriptsize $\pm 0.81$}  & \textbf{16.44} {\scriptsize $\pm 0.42$} & \textbf{16.85} {\scriptsize $\pm 0.42$}   & 5.25         \\
		\bottomrule
	\end{tabular}  
    }\\[4pt] 
    \begin{flushleft}
    \footnotesize 
    \textit{Note:} `Failed' denotes failing to produce valid results, caused by either component misidentification (e.g. sparse regression) or non-convergence (e.g. Endnet).
    \end{flushleft}
    \label{tab_mainresults}
\end{table*}

\subsection{Datasets and evaluation metrics}

We established two single-channel unmixing datasets, \textit{RRUFF-2Mix} and \textit{UNIPR-2Mix}, derived from RRUFF~\cite{LafuenteDownsYangStone+2016+1+30} and UNIPR~\cite{liu2017deep} databases, respectively. All spectra were standardized to a length of 1,024. \textit{RRUFF-2Mix} comprises 60k/5k/5k samples for training/validation/testing, while the smaller \textit{UNIPR-2Mix} contains 5k/1k/1k samples.

To generate synthetic mixtures, we randomly selected two pure spectra $\mathbf{s_1}$ and $\mathbf{s_2}$, and a mixing factor $\alpha \in [0, 1]$ to form clean mixed spectrum $\mathbf{m} = \alpha \mathbf{s_1} + (1 - \alpha) \mathbf{s_2}$. To simulate real-world conditions where spectra are contaminated by sample matrix effects and detector fluctuations, we applied a robust noise protocol. Following \cite{ZHAO2022103441, refId0, https://doi.org/10.1002/jbio.202100379}, we added a mixture of Brown (red) and white noise at a 1:1 ratio to the clean spectra, with a Signal-to-Noise Ratio (SNR) varying between 10 and 20 dB.

To verify generalization, we acquired 21 real-world mixed spectra, covering a wide range of solid-solid, solid-liquid, and liquid-liquid mixtures. For the solid samples demonstrated in Fig. \ref{fig_realworld}, we purchased powders of minerals such as Orpiment and Realgar. We first obtained pure spectral signatures, then physically mixed the powders, compressed them into pellets, and performed measurements. All real-world spectra were acquired using an 830nm laser transmission Raman spectrometer (Smekal, TRS3) and a backscatter probe.

In terms of evaluation metrics, we reported both SI-SNR and SI-SNR improvement (SI-SNRi)~\cite{8683855} for speech separation methods, and only SI-SNR for hyperspectral unmixing methods, as SI-SNRi is not applicable. For both evaluation metrics, higher values indicate better separation performance.

\begin{figure*}[tb]
  \centering
  \includegraphics[width=0.8\linewidth]{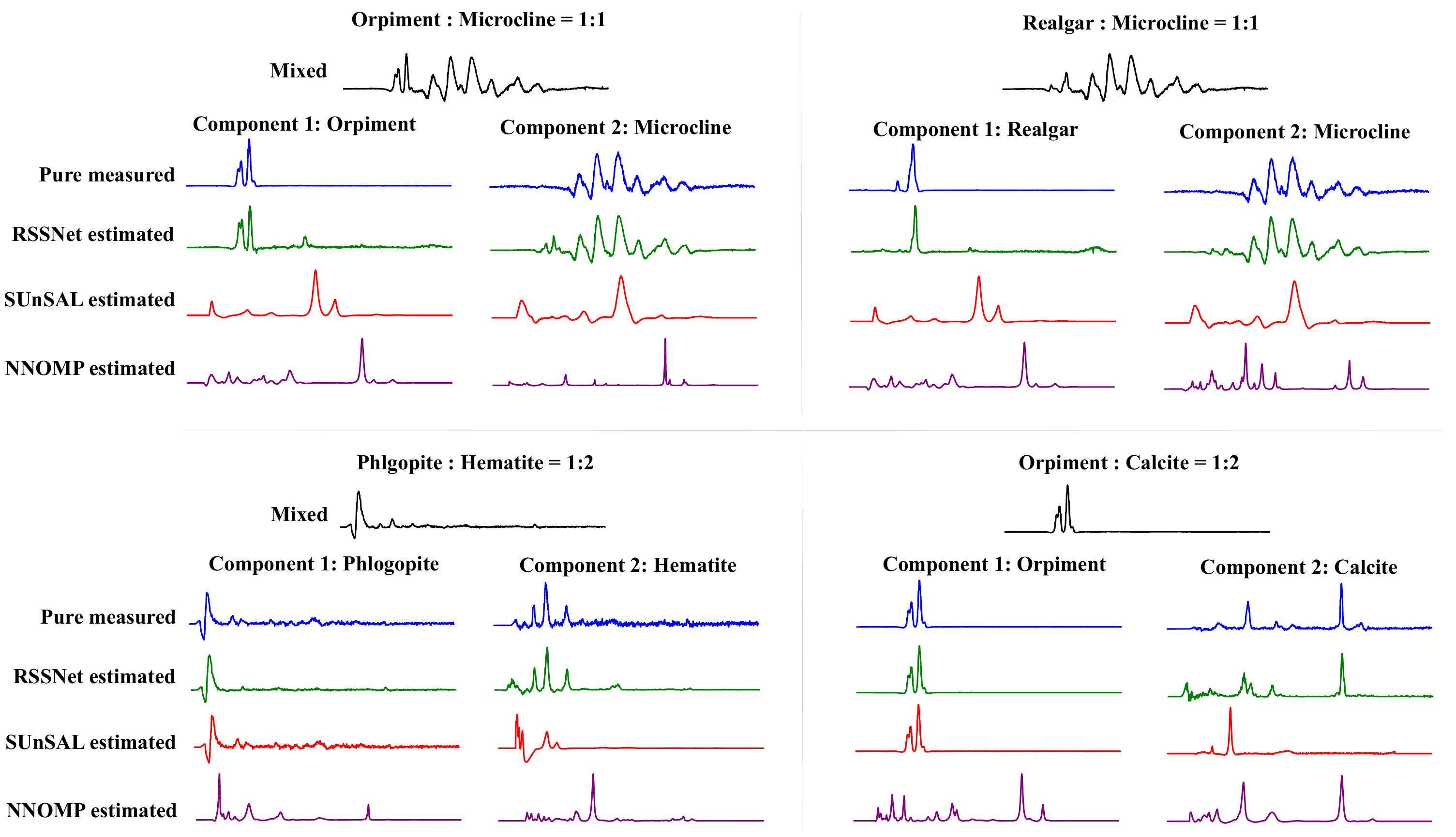}
  \caption{{\small Results of our RSSNet and existing methods unmixing real-world Raman spectra of mixtures of mineral powders Orpiment [\ce{As2S3}], Microcline [\ce{K(AlSi3O8)}], Realgar [\ce{AsS}], Hematite [\ce{Fe2O3}], Phlogopite [\ce{KMg3(AlSi3O10)(OH)2}] and Calcite [\ce{Ca(CO3)}]. The intensities of the pure spectra in the two mixed spectra above are comparable, and the characteristic peaks of both pure spectra are clearly visible in the mixed spectra. In contrast, the pure spectra in the two mixed spectra below differ significantly in magnitude; in the mixed spectra, only the peaks of the large one are clearly observable, while the other exhibits much lower amplitude and is nearly invisible. For visualization purposes, all spectra have been normalized. Please view at full resolution for better clarity.} }
    \label{fig_realworld}
\end{figure*}

\begin{table*}[tb]
  \centering
  \caption{Comparison of neural separation-based methods on extended metrics. For all metrics, lower values are better (↓).}
  \label{tab:shape_metrics}
  \begin{tabular}{lcccccc}
    \toprule
    & \multicolumn{3}{c}{\textbf{RRUFF-2Mix}} & \multicolumn{3}{c}{\textbf{UNIPR-2Mix}} \\
    \cmidrule(lr){2-4} \cmidrule(lr){5-7}
    \textbf{Method} & \textbf{SID (↓)} & \textbf{SAD (↓)} & \textbf{RMSE (↓)} & \textbf{SID (↓)} & \textbf{SAD (↓)} & \textbf{RMSE (↓)} \\
    \midrule
    AFRCNN      & 0.2275 & 0.2222 & 0.0068 & 0.3394 & 0.3084 & 0.0094 \\
    Conv-TasNet & 0.2916 & 0.2992 & 0.0091 & 0.4405 & 0.3232 & 0.0098 \\
    DPRNN       & 0.1897 & 0.2175 & 0.0066 & 0.3286 & 0.3033 & 0.0092 \\
    TDANet      & 0.3368 & 0.3267 & 0.0099 & 0.4176 & 0.3798 & 0.0115 \\
    \textbf{RSSNet (ours)} & \textbf{0.1256} & \textbf{0.1432} & \textbf{0.0043} & \textbf{0.2351} & \textbf{0.2320} & \textbf{0.0070} \\
    \bottomrule
  \end{tabular}
\end{table*}

\subsection{Compared methods}

To justify the need for our proposed RSSNet, we validate all existing unmixing paradigms and show that none of them are applicable to our case as shown in Table~\ref{tab_mainresults}. 
Note that as discussed in the introduction, all existing unmixing methods except for sparse regression are designed to work with multiple mixed spectra, so we transformed our single-channel unmixing problem to multiple channel unmixing by stacking all the spectra to form an "hyperspectral image", so that these methods can be evaluated. This is only for the purpose of evaluation mathematically, and inapplicable in practice. Specifically, for geometrical and statistical methods (e.g., VCA, FastICA), we upsampled the spectral length to 2,048 for the \textit{RRUFF-2Mix} dataset to satisfy the requirement that the spectral dimension must exceed the number of endmembers ($L > P$). 

To demonstrate the superiority of the network architecture of RSSNet, we compared RSSNet with state-of-the-art speech separation networks Conv-TasNet~\cite{ConvTasNet}, DPRNN~\cite{DPRNN2020}, AFRCNN~\cite{AFRCNN2021}, TDANet~\cite{TDANet2023} where we draw inspiration from. 

\subsection{Accuracy and robustness}

\textbf{On synthetic datasets.}
As detailed in Table~\ref{tab_mainresults}, RSSNet achieves state-of-the-art performance, surpassing baselines with an SI-SNRi of 22.50 dB (\textit{RRUFF-2Mix}) and 16.85 dB (\textit{UNIPR-2Mix}) using only 5.25M parameters. In stark contrast, traditional paradigms proved ineffective for this single-channel task. Sparse regression methods failed to identify the correct components from the spectral library, often selecting incorrect dictionary atoms due to their sensitivity to noise. This resulted in invalid reconstructed spectra as shown in Fig.~\ref{fig_realworld}. Meanwhile, geometrical and hybrid approaches produced unusable results (often negative SI-SNR) as they are fundamentally designed to require multiple mixed spectra. Also, we further evaluate shape recovery using Spectral Information Divergence (SID), Spectral Angle Distance (SAD), and RMSE. As shown in Table III, RSSNet outperforms baselines across all metrics on both datasets. This confirms RSSNet's ability to reconstruct precise spectral shapes while maintaining high noise separation performance—a key objective in Raman unmixing.

\textbf{On real-world data.}
Qualitative results in Fig.~\ref{fig_realworld} demonstrate that RSSNet accurately recovers pure spectral signatures, even in extreme scenarios where one component is spectrally submerged by the other. In contrast, existing methods SUnSAL and NNOMP both failed to unmix all four mixed spectra. In particular, SUnSAL outputted the same unmixed spectra for two different mixtures. For the other two mixtures, SUnSAL managed to find one correct component which dominated the mixed spectra, and it completely missed the other component almost invisible in the mixed spectra. 

While Fig.~\ref{fig_realworld} demonstrates qualitative feasibility on mineral powders, we conducted a rigorous quantitative evaluation on a larger dataset of 21 physically mixed samples. This dataset includes diverse scenarios such as solid-solid mixtures (e.g., Orpiment-Realgar), liquid-liquid mixtures (e.g., Ethanol-Methanol), and solid-liquid compositions. 
As shown in Table~\ref{tab:realworld_summary}, RSSNet demonstrated superior sim-to-real generalization, achieving a Mean SI-SNR of 11.72 dB (Median: 11.22 dB) and securing the best performance in 16 out of 21 cases. In stark contrast, generic speech separation models struggled to bridge the domain gap: Conv-TasNet dropped to 6.31 dB, while DPRNN failed to yield meaningful results (0.09 dB mean). This highlights that our specialized architecture is essential for bridging the sim-to-real gap and handling the complex baselines and noise in physical spectra.

\begin{figure}[t!]
  \centering
  \subfloat{
  \includegraphics[width=0.9\linewidth]{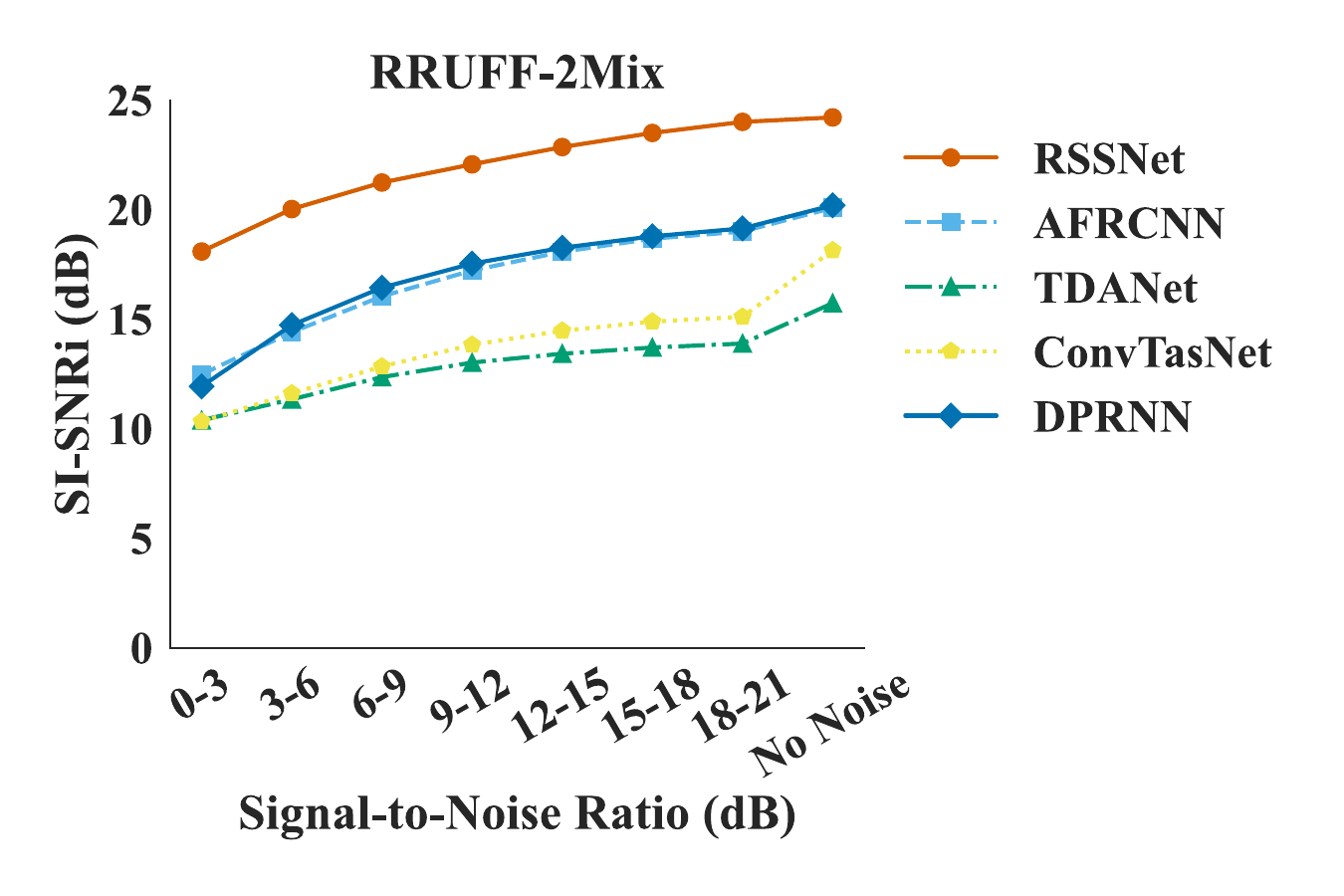}
  }
  \\
  \subfloat{
  \includegraphics[width=0.9\linewidth]{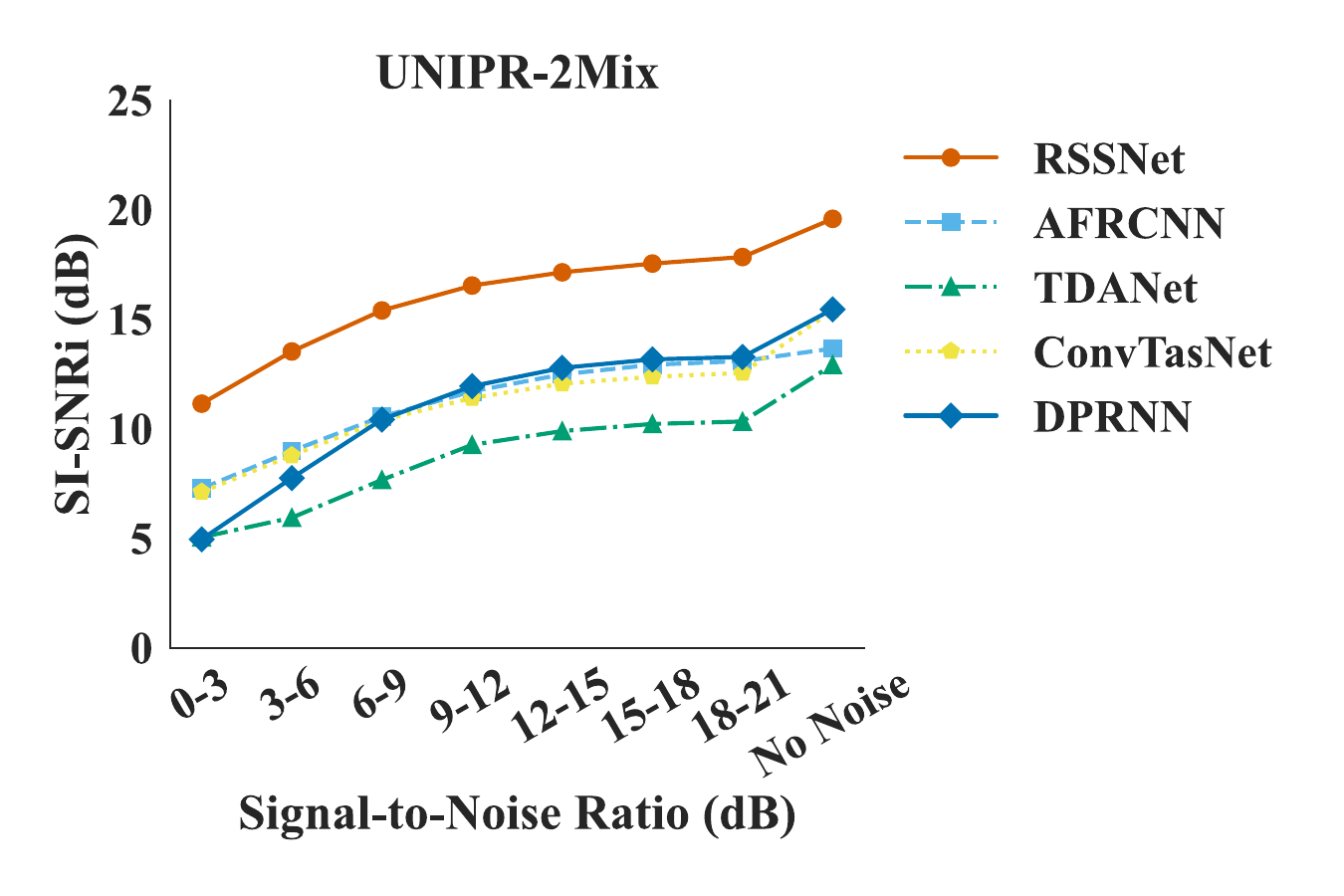}
  }
  \caption{{\small Robustness against different levels of noises of RSSNet and competing methods.} }
    \label{fig_robustness}
\end{figure}

\textbf{Robustness.} 
We conducted experiments to investigate the performance of compared methods at different level of noises (Fig.~\ref{fig_robustness}). It is evident that our RSSNet significantly outperformed the competing networks in speech separation.

\begin{table}[t!]
    \centering
    \caption{Summary of Quantitative SI-SNR results on 21 real-world mixed samples. Higher values indicate better performance.}
    \label{tab:realworld_summary}
    \renewcommand{\arraystretch}{1.1}
    \setlength{\tabcolsep}{10pt}
    
    \begin{tabular}{lcc}
        \toprule
        \textbf{Method} & \textbf{Mean} (dB) & \textbf{Median} (dB) \\
        \midrule
        ConvTasNet & 6.31 & 5.51 \\
        DPRNN & 0.09 & 2.31 \\
        AFRCNN & 9.32 & 10.38 \\
        TDANet & 8.81 & 7.28 \\
        \textbf{RSSNet (ours)} & \textbf{11.72} & \textbf{11.22} \\
        \bottomrule
    \end{tabular}
\end{table}

\begin{table}[t!]
\caption{Ablation studies on RSSNet components (RRUFF-2Mix). Left: Encoder kernel size. Middle: DWConv path position (see Fig. \ref{fig:2}). Right: DWConv kernel size.}
\centering
\renewcommand{\arraystretch}{1.1}
\begin{tabular}{cc|cc|cc}
\hline
\multicolumn{2}{c|}{\textbf{Encoder Kernel}} & \multicolumn{2}{c|}{\textbf{DWConv Path}} & \multicolumn{2}{c}{\textbf{DWConv Kernel}} \\
Size & SI-SNRi (dB) & Path & SI-SNRi (dB) & Size & SI-SNRi (dB) \\
\hline
2 & 19.53 & None & 22.29 & $\mathbf{1 \times 1}$ & $\mathbf{22.50}$ \\
$\mathbf{3}$ & $\mathbf{22.50}$ & $\mathbf{p_1}$ & $\mathbf{22.50}$ & $3 \times 3$ & 19.95 \\
5 & 22.34 & $p_2$ & 20.22 & $5 \times 5$ & 20.16 \\
7 & 19.26 & $p_3$ & 20.06 & $7 \times 7$ & 18.43 \\
9 & 20.06 & - & - & - & - \\
\hline
\end{tabular}
\label{tab:ablation}
\end{table}

\subsection{Ablation study} \label{sec:ablation}

To validate the architectural design of RSSNet, we conducted ablation studies on the \textit{RRUFF-2Mix} dataset, focusing on the Raman encoder's kernel size and the configuration of the depth-wise convolution (\textit{DWConv}) in the separator.

\textbf{Raman encoder kernel size.}
Raman spectra are characterized by sharp, distinct peaks where position and shape are critical. We hypothesized that a smaller kernel size is preferable to capture these fine-scale features. As shown in Table \ref{tab:ablation}, the network achieved the best performance (22.50 dB SI-SNRi) with a kernel size of 3. Performance gradually declined as the kernel size increased (dropping to 20.06 dB at size 9), confirming that extracting high-frequency local details is more important than a large initial receptive field for this task.

\textbf{Depth-wise convolution strategy.}
We further investigated the role of the \textit{DWConv} layer within the RSSNet Block. We evaluated three potential paths for this layer ($p_1, p_2, p_3$) as illustrated in Fig. \ref{fig:2}. Placing the \textit{DWConv} on path $p_1$ yielded the highest accuracy. Furthermore, we examined the kernel size for this specific layer. Results indicate that a $1 \times 1$ kernel significantly outperforms larger ones (e.g., 22.50 dB vs. 18.43 dB for $7 \times 7$). This $1 \times 1$ \textit{DWConv} acts as a channel-mixing mechanism similar to Point-wise Convolution or the projection in self-attention, effectively modeling global correlations among feature channels. Since the TDA modules already handle temporal dependencies, using larger kernels here is redundant and degrades local sensitivity.

\section{Conclusion and Future Work}
In this paper, we presented a novel neural separation paradigm for single-channel Raman spectral unmixing and introduce RSSNet, a brain-inspired neural network successfully overcomes the limitations of traditional geometrical and statistical methods in underdetermined and noisy scenarios. Experimental results on both synthetic datasets and real-world mineral samples confirm that RSSNet achieves state-of-the-art performance in decomposing single mixed spectra, demonstrating strong robustness against spectral variations. Future work will focus on extending the framework to handle multi-component mixtures and investigating generalization to unseen substances in open-world scenarios.

\bibliographystyle{IEEEtran}
\bibliography{rssnet}

\begin{thebibliography}{10}
\providecommand{\url}[1]{#1}
\csname url@samestyle\endcsname
\providecommand{\newblock}{\relax}
\providecommand{\bibinfo}[2]{#2}
\providecommand{\BIBentrySTDinterwordspacing}{\spaceskip=0pt\relax}
\providecommand{\BIBentryALTinterwordstretchfactor}{4}
\providecommand{\BIBentryALTinterwordspacing}{\spaceskip=\fontdimen2\font plus
\BIBentryALTinterwordstretchfactor\fontdimen3\font minus
  \fontdimen4\font\relax}
\providecommand{\BIBforeignlanguage}[2]{{%
\expandafter\ifx\csname l@#1\endcsname\relax
\typeout{** WARNING: IEEEtran.bst: No hyphenation pattern has been}%
\typeout{** loaded for the language `#1'. Using the pattern for}%
\typeout{** the default language instead.}%
\else
\language=\csname l@#1\endcsname
\fi
#2}}
\providecommand{\BIBdecl}{\relax}
\BIBdecl

\bibitem{liu2017deep}
J.~Liu, M.~Osadchy, L.~Ashton, M.~Foster, C.~J. Solomon, and S.~J. Gibson,
  ``Deep convolutional neural networks for raman spectrum recognition: a
  unified solution,'' \emph{Analyst}, vol. 142, no.~21, pp. 4067--4074, 2017.

\bibitem{Sunsal2010}
J.~M. Bioucas-Dias and M.~A.~T. Figueiredo, ``Alternating direction algorithms
  for constrained sparse regression: Application to hyperspectral unmixing,''
  in \emph{2010 2nd Workshop on Hyperspectral Image and Signal Processing:
  Evolution in Remote Sensing}, 2010, pp. 1--4.

\bibitem{SunsalLegacy2021}
M.~Parente and M.-D. Iordache, ``Sparse unmixing of hyperspectral data: The
  legacy of sunsal,'' in \emph{2021 IEEE International Geoscience and Remote
  Sensing Symposium IGARSS}, 2021, pp. 21--24.

\bibitem{FLCS2001}
D.~Heinz and Chein-I-Chang, ``Fully constrained least squares linear spectral
  mixture analysis method for material quantification in hyperspectral
  imagery,'' \emph{IEEE Transactions on Geoscience and Remote Sensing},
  vol.~39, no.~3, pp. 529--545, 2001.

\bibitem{garrido2008multivariate}
M.~Garrido, F.~Rius, and M.~Larrechi, ``Multivariate curve
  resolution--alternating least squares (mcr-als) applied to spectroscopic data
  from monitoring chemical reactions processes,'' \emph{Analytical and
  bioanalytical chemistry}, vol. 390, no.~8, pp. 2059--2066, 2008.

\bibitem{kouakou2024fly}
H.~Kouakou, J.~H. de~Morais~Goulart, R.~Vitale, T.~Oberlin, D.~Rousseau,
  C.~Ruckebusch, and N.~Dobigeon, ``On-the-fly spectral unmixing based on
  kalman filtering,'' \emph{Chemometrics and Intelligent Laboratory Systems},
  p. 105252, 2024.

\bibitem{UnmixingAE2024PNAS}
D.~Georgiev, Álvaro Fernández-Galiana, S.~V. Pedersen, G.~Papadopoulos,
  R.~Xie, M.~M. Stevens, and M.~Barahona, ``Hyperspectral unmixing for raman
  spectroscopy via physics-constrained autoencoders,'' \emph{Proceedings of the
  National Academy of Sciences}, vol. 121, no.~45, p. e2407439121, 2024.

\bibitem{mDAE}
R.~Guo, W.~Wang, and H.~Qi, ``Hyperspectral image unmixing using autoencoder
  cascade,'' in \emph{2015 7th Workshop on Hyperspectral Image and Signal
  Processing: Evolution in Remote Sensing (WHISPERS)}, 2015, pp. 1--4.

\bibitem{SIDAEU}
F.~Palsson, J.~Sigurdsson, J.~R. Sveinsson, and M.~O. Ulfarsson, ``Neural
  network hyperspectral unmixing with spectral information divergence
  objective,'' in \emph{2017 IEEE International Geoscience and Remote Sensing
  Symposium (IGARSS)}, 2017, pp. 755--758.

\bibitem{Endnet}
S.~Ozkan, B.~Kaya, and G.~B. Akar, ``Endnet: Sparse autoencoder network for
  endmember extraction and hyperspectral unmixing,'' \emph{IEEE Transactions on
  Geoscience and Remote Sensing}, vol.~57, no.~1, pp. 482--496, 2019.

\bibitem{CNNAEU}
B.~Palsson, M.~O. Ulfarsson, and J.~R. Sveinsson, ``Convolutional autoencoder
  for spectral–spatial hyperspectral unmixing,'' \emph{IEEE Transactions on
  Geoscience and Remote Sensing}, vol.~59, no.~1, pp. 535--549, 2021.

\bibitem{5585746}
X.~Liu, W.~Xia, B.~Wang, and L.~Zhang, ``An approach based on constrained
  nonnegative matrix factorization to unmix hyperspectral data,'' \emph{IEEE
  Transactions on Geoscience and Remote Sensing}, vol.~49, no.~2, pp. 757--772,
  2011.

\bibitem{VCA2005}
J.~Nascimento and J.~Dias, ``Vertex component analysis: a fast algorithm to
  unmix hyperspectral data,'' \emph{IEEE Transactions on Geoscience and Remote
  Sensing}, vol.~43, no.~4, pp. 898--910, 2005.

\bibitem{Li2017Sparse}
C.~Li, Y.~Ma, X.~Mei, F.~Fan, J.~Huang, and J.~Ma, ``Sparse unmixing of
  hyperspectral data with noise level estimation,'' \emph{Remote Sensing},
  vol.~9, no.~11, 2017.

\bibitem{XU201846}
X.~Xu, Z.~Shi, and B.~Pan, ``$\ell_0$-based sparse hyperspectral unmixing using
  spectral information and a multi-objectives formulation,'' \emph{ISPRS
  Journal of Photogrammetry and Remote Sensing}, vol. 141, pp. 46--58, 2018.

\bibitem{Gong2017Multiobjective}
M.~Gong, H.~Li, E.~Luo, J.~Liu, and J.~Liu, ``A multiobjective cooperative
  coevolutionary algorithm for hyperspectral sparse unmixing,'' \emph{IEEE
  Transactions on Evolutionary Computation}, vol.~21, no.~2, pp. 234--248,
  2017.

\bibitem{Drumetz2016Blind}
L.~Drumetz, M.-A. Veganzones, S.~Henrot, R.~Phlypo, J.~Chanussot, and
  C.~Jutten, ``Blind hyperspectral unmixing using an extended linear mixing
  model to address spectral variability,'' \emph{IEEE Transactions on Image
  Processing}, vol.~25, no.~8, pp. 3890--3905, 2016.

\bibitem{8369155}
D.~Wang and J.~Chen, ``Supervised speech separation based on deep learning: An
  overview,'' \emph{IEEE/ACM Transactions on Audio, Speech, and Language
  Processing}, vol.~26, no.~10, pp. 1702--1726, 2018.

\bibitem{ConvTasNet}
Y.~Luo and N.~Mesgarani, ``Conv-tasnet: Surpassing ideal time–frequency
  magnitude masking for speech separation,'' \emph{IEEE/ACM Transactions on
  Audio, Speech, and Language Processing}, vol.~27, no.~8, pp. 1256--1266,
  2019.

\bibitem{DPRNN2020}
Y.~Luo, Z.~Chen, and T.~Yoshioka, ``Dual-path rnn: Efficient long sequence
  modeling for time-domain single-channel speech separation,'' in \emph{ICASSP
  2020 - 2020 IEEE International Conference on Acoustics, Speech and Signal
  Processing (ICASSP)}, 2020, pp. 46--50.

\bibitem{TDANet2023}
K.~Li, R.~Yang, and X.~Hu, ``An efficient encoder-decoder architecture with
  top-down attention for speech separation,'' in \emph{The Eleventh
  International Conference on Learning Representations}, 2023.

\bibitem{DeepClustering}
J.~R. Hershey, Z.~Chen, J.~Le~Roux, and S.~Watanabe, ``Deep clustering:
  Discriminative embeddings for segmentation and separation,'' in \emph{2016
  IEEE International Conference on Acoustics, Speech and Signal Processing
  (ICASSP)}, 2016, pp. 31--35.

\bibitem{TasNet}
Y.~Luo and N.~Mesgarani, ``Tasnet: Time-domain audio separation network for
  real-time, single-channel speech separation,'' in \emph{2018 IEEE
  International Conference on Acoustics, Speech and Signal Processing
  (ICASSP)}, 2018, pp. 696--700.

\bibitem{AFRCNN2021}
X.~Hu, K.~Li, W.~Zhang, Y.~Luo, J.-M. Lemercier, and T.~Gerkmann, ``Speech
  separation using an asynchronous fully recurrent convolutional neural
  network,'' in \emph{Advances in Neural Information Processing Systems},
  M.~Ranzato, A.~Beygelzimer, Y.~Dauphin, P.~Liang, and J.~W. Vaughan, Eds.,
  vol.~34, 2021, pp. 22\,509--22\,522.

\bibitem{donghao2024moderntcn}
D.~Luo and X.~Wang, ``Modern{TCN}: A modern pure convolution structure for
  general time series analysis,'' in \emph{The Twelfth International Conference
  on Learning Representations}, 2024.

\bibitem{vaswani2017attention}
A.~Vaswani, N.~Shazeer, N.~Parmar, J.~Uszkoreit, L.~Jones, A.~N. Gomez, L.~u.
  Kaiser, and I.~Polosukhin, ``Attention is all you need,'' in \emph{Advances
  in Neural Information Processing Systems}, I.~Guyon, U.~V. Luxburg,
  S.~Bengio, H.~Wallach, R.~Fergus, S.~Vishwanathan, and R.~Garnett, Eds.,
  vol.~30, 2017.

\bibitem{NNOMP_Unmixing}
M.~Yaghoobi, D.~Wu, R.~J. Clewes, and M.~E. Davies, ``{Fast sparse Raman
  spectral unmixing for chemical fingerprinting and quantification},'' in
  \emph{Optics and Photonics for Counterterrorism, Crime Fighting, and Defence
  XII}, D.~Burgess, G.~Owen, H.~Bouma, F.~Carlysle-Davies, R.~J. Stokes, and
  Y.~Yitzhaky, Eds., vol. 9995, International Society for Optics and
  Photonics.\hskip 1em plus 0.5em minus 0.4em\relax SPIE, 2016, p. 99950E.

\bibitem{FastICA1997}
A.~Hyvärinen and E.~Oja, ``A fast fixed-point algorithm for independent
  component analysis,'' \emph{Neural Computation}, vol.~9, no.~7, pp.
  1483--1492, 07 1997.

\bibitem{NMF1999}
D.~D. Lee and H.~S. Seung, ``Learning the parts of objects by non-negative
  matrix factorization,'' \emph{Nature}, vol. 401, no. 6755, pp. 788--791, Oct.
  1999.

\bibitem{NFINDR1999}
M.~E. Winter, ``N-findr: an algorithm for fast autonomous spectral end-member
  determination in hyperspectral data,'' in \emph{SPIE Proceedings}, vol.
  3753.\hskip 1em plus 0.5em minus 0.4em\relax SPIE, October 1999, pp.
  266--275.

\bibitem{MCRALSraman2022}
I.~Matveeva, I.~Bratchenko, Y.~Khristoforova, L.~Bratchenko, A.~Moryatov,
  S.~Kozlov, O.~Kaganov, and V.~Zakharov, ``Multivariate curve resolution
  alternating least squares analysis of in vivo skin raman spectra,''
  \emph{Sensors}, vol.~22, no.~24, 2022.

\bibitem{LafuenteDownsYangStone+2016+1+30}
B.~Lafuente, R.~T. Downs, H.~Yang, and N.~Stone, \emph{1. The power of
  databases: The RRUFF project}, Berlin, München, Boston, 2016, pp. 1--30.

\bibitem{ZHAO2022103441}
Y.~Zhao, G.~Che, and X.~Zhao, ``Adaptive noise removal for biological raman
  spectra with low snr,'' \emph{Vibrational Spectroscopy}, vol. 123, p. 103441,
  2022.

\bibitem{refId0}
{Moester, M. J. B.}, {Ariese, F.}, and {de Boer, J. F.}, ``Optimized
  signal-to-noise ratio with shot noise limited detection in stimulated raman
  scattering microscopy,'' \emph{J. Eur. Opt. Soc.-Rapid Publ.}, vol.~10, p.
  15022, 2015.

\bibitem{https://doi.org/10.1002/jbio.202100379}
R.~Ranjan, G.~Costa, M.~A. Ferrara, M.~Sansone, and L.~Sirleto, ``Noises
  investigations and image denoising in femtosecond stimulated raman scattering
  microscopy,'' \emph{Journal of Biophotonics}, vol.~15, no.~6, p. e202100379,
  2022.

\bibitem{8683855}
J.~L. Roux, S.~Wisdom, H.~Erdogan, and J.~R. Hershey, ``Sdr – half-baked or
  well done?'' in \emph{ICASSP 2019 - 2019 IEEE International Conference on
  Acoustics, Speech and Signal Processing (ICASSP)}, 2019, pp. 626--630.

\end{thebibliography}

\end{document}